\documentclass[journal]{IEEEtran}
\usepackage{graphicx,float,amssymb,multirow,pifont,amsmath,scalerel,makecell}
\usepackage{bm,hyperref,xcolor,enumitem,booktabs,arydshln,footnote,url,threeparttable}
\usepackage[caption=false]{subfig}
\usepackage{color}
\usepackage{cite}
\usepackage[numbers,sort&compress]{natbib}
\usepackage{amsthm,amsmath,amssymb}
\usepackage{mathrsfs}
\usepackage{threeparttable}

\usepackage[subfigure,titles]{tocloft}
\hypersetup{hypertex=true,
	colorlinks=true,
	linkcolor=blue,
	anchorcolor=blue,
	citecolor=blue}

\begin{document}
\title{RS3Mamba: Visual State Space Model for Remote Sensing Images Semantic Segmentation}
\author{Xianping~Ma,
	Xiaokang~Zhang,~\IEEEmembership{Member,~IEEE,}
	and~Man-On~Pun,~\IEEEmembership{Senior Member,~IEEE}
	\thanks{This work was supported in part by the Guangdong Provincial Key Laboratory of Future Networks of Intelligence under Grant 2022B1212010001 and National Natural Science Foundation of China under Grant 42371374 and 41801323. \textit{(Corresponding author: Man-On Pun; Xiaokang Zhang)}}
	\thanks{Xianping Ma and Man-On Pun are with the School of Science and Engineering, the Future Network of Intelligence Institute (FNii), the Chinese University of Hong Kong, Shenzhen, Shenzhen 518172, China (e-mail: xianpingma@link.cuhk.edu.cn; SimonPun@cuhk.edu.cn).}
	\thanks{Xiaokang Zhang is with the School of Information Science and Engineering, Wuhan University of Science and Technology, Wuhan 430081, China (e-mail: natezhangxk@gmail.com).}}
\maketitle

\begin{abstract}
Semantic segmentation of remote sensing images is a fundamental task in geoscience research. However, there are some significant shortcomings for the widely used convolutional neural networks (CNNs) and Transformers. The former is limited by its insufficient long-range modeling capabilities, while the latter is hampered by its computational complexity. Recently, a novel visual state space (VSS) model represented by Mamba has emerged, capable of modeling long-range relationships with linear computability. In this work, we propose a novel dual-branch network named remote sensing images semantic segmentation Mamba (RS3Mamba) to incorporate this innovative technology into remote sensing tasks. Specifically, RS3Mamba utilizes VSS blocks to construct an auxiliary branch, providing additional global information to convolution-based main branch. Moreover, considering the distinct characteristics of the two branches, we introduce a collaborative completion module (CCM) to enhance and fuse features from the dual-encoder. Experimental results on two widely used datasets, ISPRS Vaihingen and LoveDA Urban, demonstrate the effectiveness and potential of the proposed RS3Mamba. To the best of our knowledge, this is the first vision Mamba specifically designed for remote sensing images semantic segmentation. The source code will be made available at \href{https://github.com/sstary/SSRS}{https://github.com/sstary/SSRS}.
\end{abstract}


\IEEEpeerreviewmaketitle

\section{Introduction}\label{sec:int}
Modern geoscience research heavily relies on a wide range of remote sensing data collected by satellite or aerial devices. These data play a crucial role in capturing the spectral characteristics of objects on Earth's surface and offering accurate visual representations of both natural and human-made structures. Semantic segmentation methods aim to classify every pixel in remote sensing images into distinct categories, thereby assisting researchers in thoroughly exploring surface conditions. This automated analysis and interpretation approach supports various downstream tasks and applications, including land cover mapping, environmental monitoring, and disaster management. 

In recent years, deep learning-based methods as a data-driven automated technology have revolutionized semantic segmentation methods. Compared with conventional methods based on expert knowledge and artificial designation, deep learning has the capability to autonomously extract effective features from data and generate predicted probabilities in an end-to-end manner. Currently, mainstream models in remote sensing can be categorized into two main types: CNNs and Transformers. The former can extract image features through stacked convolution operations while the latter can model long-range dependencies based on the self-attention mechanism. Nowadays, these methods have inspired a large number of remote sensing images processing models, including pure convolutional network \cite{diakogiannis2020resunet}, transformer-based network \cite{xu2021efficient} and hybrid architecture \cite{ma2024miltulevel}.

However, despite their advantages, these models have their limitations when applied to remote sensing images. Compared with natural images, remote sensing images has the characteristics of complex scenes and significant variations in object scales. As a result, CNNs are constrained by their local receptive fields, making it challenging to grasp and learn the intricate representations. On the other hand, although Transformers possess the capability to learn long-range dependencies, their high computational complexity poses a significant challenge when considering model efficiency and memory footprint. Recently, Mamba \citep{gu2023mamba}, built upon the state space model (SSM) \citep{gu2021efficiently}, has emerged as an alternative for establishing long-distance dependency relationships while maintaining linear computational complexity. Subsequently, Vim \citep{zhu2024vision} and VMamba \citep{liu2024vmamba} have extended this mechanism into the field of image processing. In remote sensing, Pan-Mamba \citep{he2024pan} proposed channel swapping Mamba and cross-modal Mamba for pan-sharpening. RSMamba \citep{chen2024rsmamba} presented a multi-path VSS Block for large-scale image interpretation. These two methods directly replace the existing network with the VSS blocks. However, since most existing VSS-based models are trained from scratch, they are temporarily hard to compare with fully pre-trained CNNs and Transformers in terms of optimal performance.

To introduce the VSS module into remote sensing images semantic segmentation and cope with the aforementioned challenge, we propose an auxiliary branch strategy that utilizes VSS blocks to supply additional global information, aiding the convolution-based main branch in feature extraction. Furthermore, to address the disparity between global and local semantics, we introduce a CCM module to facilitate cross-branch semantic fusion. The contributions of this work can be outlined as follows:
\begin{itemize}
\item We propose RS3Mamba, marking the first exploration of the potential application of VSS-based models in remote sensing images semantic segmentation. It provides valuable insights for the future development of more efficient and effective VSS-based methods for remote sensing tasks.

\item We propose a novel CCM for feature fusion, which can bridge the features from the dual-branch encoder, making them more effective in representation learning of remote sensing images.

\item Extensive experiments on two well-known, publicly available remote sensing datasets, ISPRS Vaihingen and LoveDA Urban confirm that the proposed method holds significant advantages over existing methods based on CNNs and Transformers.
\end{itemize}

\begin{figure}[t]
\centering
{\includegraphics[width=0.95\linewidth]{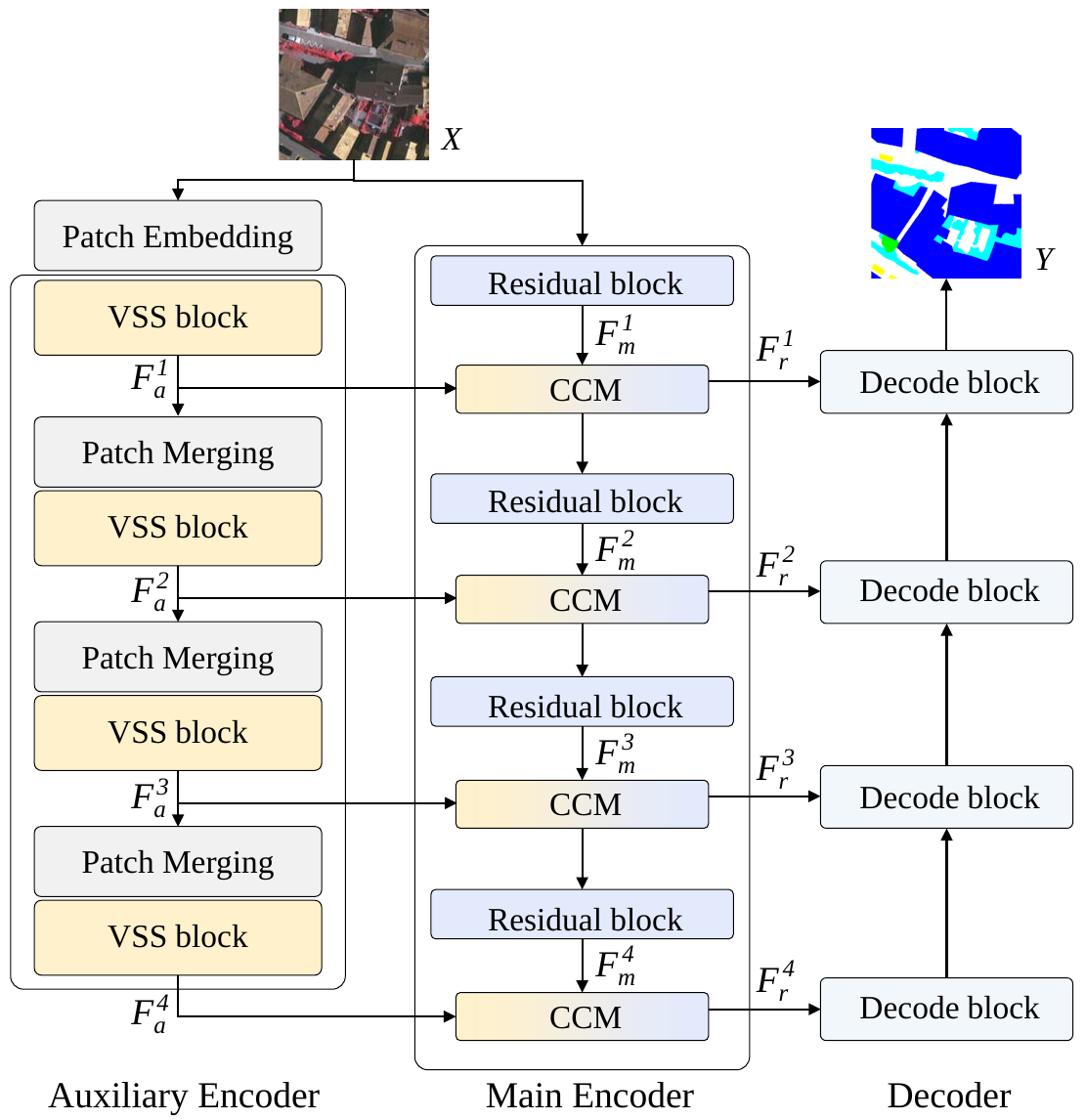}}
\caption{The overall architecture of RS3Mamba.}
\label{fig1}
\end{figure}

\section{Methodology}\label{sec:met}
The RS3Mamba consists of three parts, namely the VSS auxiliary encoder, the residual main encoder including the CCM for cross-branch semantic fusion, and the decoder as depicted in Fig.~\ref{fig1}. Specifically, the RS3Mamba model extracts image features via the main and auxiliary branches, each comprising four corresponding blocks. The features generated by the auxiliary branch are fed into the CCM of the corresponding scale within the main branch for feature fusion. After the feature extraction and feature fusion at four scales, multiscale features are obtained before they are fed into the decoder with skip connections to generate the prediction map. The decoder in UNetformer \cite{wang2022unetformer} is adopted in this work.

\subsection{Auxiliary encoder}
\begin{figure}[t]
\centering
{\includegraphics[width=1.0\linewidth]{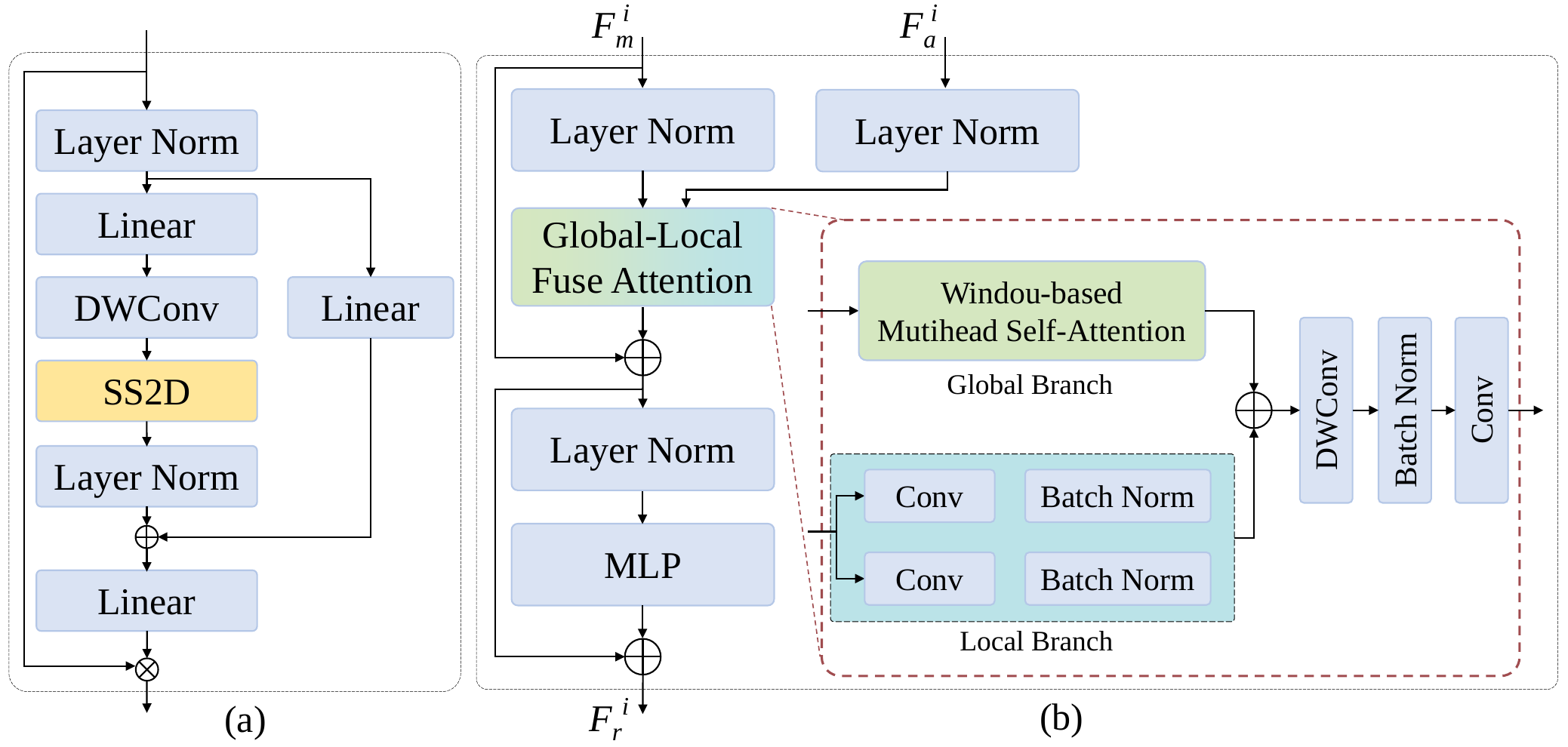}}
\caption{(a) The detailed architecture of VSS block. (b) The detailed architecture of CCM.}
\label{fig2}
\end{figure}

The auxiliary encoder is constructed based on VSS blocks that models long-range dependencies \citep{liu2024swin}. The detailed structure is presented in Fig.~\ref{fig2}(a), where 2D-selective-scan (SS2D) is the core computing unit of the VSS block. The SS2D method extends image patches in four directions to create four separate sequences. These sequences are then individually processed through the SSM. Finally, the resulting features are combined to generate the complete $2$D feature map. Given the input feature map $x$, the output feature map $\overline{x}$ of SS2D can be formulated as:
\begin{eqnarray}
x_{v} = expand(x,v),\label{eq1}\\
\overline{x}_{v} = S6(x_{v}),\label{eq2}\\
\overline{x} = merge(\overline{x}_{1},\overline{x}_{2},\overline{x}_{3},\overline{x}_{4}),\label{eq3}
\end{eqnarray}
where $v\in V = \{1,2,3,4\}$ denotes four different scanning directions. $expand(\cdot)$ and $merge(\cdot)$ denotes the $scan~expand$ and $scan~merge$ operations in \citep{liu2024vmamba}. The selective scan space state sequential model ($S6$) described in Eq.~\eqref{eq2} serves as the core operator of the VSS block. It facilitates interactions between each element in a 1D array and any of the previously scanned samples through a condensed hidden state. Please refer to \citep{liu2024vmamba} for more details about $S6$.

As depicted in Fig.~\ref{fig1}, the auxiliary encoder consists of four successive stages, each consisting of a patch operator and a VSS block. The first stage contains one Patch Embedding layer followed by the VSS block, whereas the following three stages are equipped with one Patch Merging layer followed by the VSS block each. Denote by $X\in \mathbb{R}^{{H\times W \times Z}}$ the input images. $Z$ denotes the number of image channels, whereas $H$ and $W$ represent the height and width of the image, respectively. $X$ are first split into non-overlapping patches by the Patch Embedding layer before extracting features by VSS block whose output is denoted by $F_{a}^{\emph {1}}$. The following three stages perform similar operations and generate $F_{a}^{\emph {2-4}}$. Notably, since the feature flow of the auxiliary branch are not affected by the main branch, we can perform the entire auxiliary branch firstly.

\begin{table*}[t]\footnotesize
	\centering
	\caption{Experimental results on the ISPRS Vaihingen dataset. We present the OA of five foreground classes and three overall performance metrics.  The accuracy of each category is presented in the F1/IoU form. Bold values are the best.}
	  \renewcommand\arraystretch{1.2}
		\begin{tabular}{ccccccccc}
			\hline
			\textbf{Method} & \textbf{Backbone} &  \textbf{impervious surface} & \textbf{building} & \textbf{low vegetation}  & \textbf{tree}  & \textbf{car}  & \textbf{mF1}    & \textbf{mIoU} \\
			\hline
			ABCNet  \cite{li2021abcnet}  &ResNet-18& 89.78/81.45 & 94.30/89.21 & 78.49/64.59 & 90.08/81.95 & 74.05/58.80 & 85.34 & 75.20 \\
			TransUNet \cite{chen2021transunet}   &R50-ViT-B& 90.77/83.10 & 94.32/89.25 & 79.02/65.32 & 90.53/82.70 & 82.66/70.45 & 87.46 & 78.16 \\
			UNetformer \cite{wang2022unetformer} &ResNet-18& 92.33/85.76 & 96.25/92.78 & 80.47/67.33 & 90.85/83.22 & 89.35/80.75 & 89.85 & 81.97 \\
			CMTFNet    \cite{wu2023cmtfnet}   &ResNet-50& 92.53/86.09 & \textbf{96.95}/\textbf{94.09} & 79.98/66.64 & 90.22/82.19 & 89.87/81.60 & 89.91 & 82.12 \\
			\hline
			RS3Mamba    &R18-Mamba-T& \textbf{92.83}/\textbf{86.62} & 96.82/93.83 & \textbf{80.84}/\textbf{67.84} & \textbf{91.10}/\textbf{83.66} & \textbf{90.09}/\textbf{81.97} & \textbf{90.34} & \textbf{82.78} \\
			\hline
	\end{tabular}\label{tab:vlist}
\end{table*}

\subsection{Main encoder and CCM}
The ResNet18 is adopted as the main encoder to learn the local representations. As depicted in Fig.~\ref{fig1}, it contains four residual blocks and four CCMs. The four residual blocks perform convolution operations and generate multiscale features denoted by $F_{m}^{\emph {1-4}}$. In contrast to the auxiliary encoder, the main encoder leverages the capabilities of the existing pre-trained model to efficiently extract features from remote sensing images. Hereby we incorporate the features from the auxiliary branch at each scale into the main branch by CCM to compensate for its limitation in extracting global information.

The detailed structure of CCM is shown in Fig.~\ref{fig2}(b). CCM is employed to fuse the cross-branch feature maps $F_{m}^{i}$ and $F_{a}^{i}$ where $i=\{1,2,3,4\}$. Specifically, the CCM comprises two parallel branches, also namely the global branch and the local branch. The former is used to enhance $F_{m}^{i}$ from the main branch, while the latter is used to process $F_{a}^{i}$ from the auxiliary branch. Considering that the features of the main branch are obtained by convolution operations with local properties, we use windou-based multihead self-attention \cite{liu2021swin} for modeling long-range dependencies. It should be highlighted that this mechanism also maintains linear complexity. On the other hand, considering that the features of auxiliary branches are obtained by VSS with long-range properties, we use convolution to learn the details in local. Therefore, this fusion module is named the collaborative completion module in consideration of our completion of the double-branch features. At each of the four stages in the main branch, a CCM conducts feature fusion with the corresponding scale. The resulting fusion feature, denoted as $F_{r}^{i}$ is then fed into the decoder as the skip connections. The decoder will recover the abstract features and give the final prediction map. The training objective is the cross entropy loss.

\section{Experiments and Discussion}\label{sec:exp}
\subsection{Datasets}
\subsubsection{ISPRS Vaihingen} The ISPRS Vaihingen dataset consists of $16$ true orthophotos, each boasting very high-resolution at an average size of $2500\times2000$ pixels. These orthophotos are composed of three channels: Near-Infrared, Red, and Green (NIRRG), with a ground sampling distance of $9$ cm. The dataset encompasses five foreground classes, namely {\em impervious surface}, {\em building}, {\em low vegetation}, {\em tree}, {\em car}, and one background class ({\em clutter}). In our experiments, the $16$ orthophotos are divided into a training set comprising $12$ patches and a test set comprising $4$ patches. 

\subsubsection{LoveDA Urban} The LoveDA dataset offers a comprehensive collection of scenes, including Urban and Rural. In this work, we selected the LoveDA Urban scene due to its diverse distribution of ground objects. The LoveDA Urban scene comprises $1833$ high-resolution optical remote sensing images with the size of $1024\times 1024$ pixels. Each image provides three channels, namely Red, Green, and Blue (RGB), with a ground sampling distance of $30$ cm. The dataset encompasses seven landcover categories, including {\em background}, {\em building}, {\em road}, {\em water}, {\em barren}, {\em forest}, and {\em agriculture} \cite{wang2021loveda}. The $1833$ images are partitioned into two subsets, with $1156$ images allocated for training and $677$ images for testing. 

The two datasets contrast in terms of sampling resolution, ground object categories, and label accuracy. By conducting experiments on both datasets, we can demonstrate the effectiveness and superiority of the RS3Mamba.

\subsection{Experimental Setup}
We adopted UNetformer \citep{wang2022unetformer} based on ResNet18 \citep{he2016deep} as our baseline model. The proposed RS3Mamba was benchmarked against several state-of-the-art supervised methods, including ABCNet \citep{li2021abcnet}, TransUNet \citep{chen2021transunet}, UNetformer \citep{wang2022unetformer}, and CMTFNet \citep{wu2023cmtfnet}. The experiments were conducted using PyTorch on a single NVIDIA GeForce RTX $4090$ GPU equipped with $24$GB RAM. Furthermore, Stochastic gradient descent (SGD) served as the optimization algorithm for training all models. A learning rate of $0.01$, a momentum of $0.9$, a decaying coefficient of $0.0005$, and a batch size of $10$ were employed. The total epochs were set to $50$, with one test per epoch. For quantitative evaluation of the proposed approach, two widely used metrics were recorded: mean F1-score (mF1) and mean intersection over union (mIoU).

\begin{table*}[t]\footnotesize
  \centering
	\caption{Experimental results on the LoveDA Urban dataset. We present the OA of five foreground classes and three overall performance metrics.  The accuracy of each category is presented in the F1/IoU form. Bold values are the best.}
	  \tabcolsep=0.12cm
	  \renewcommand\arraystretch{1.2}
		\begin{tabular}{ccccccccccc}
			\hline
			\textbf{Method}  & \textbf{Backbone}  & \textbf{background} & \textbf{building} & \textbf{road}  & \textbf{water}  & \textbf{barren}  & \textbf{forest} & \textbf{agriculture}  & \textbf{mF1} & \textbf{mIoU}  \\
			\hline
			ABCNet  \cite{li2021abcnet}  &ResNet-18& 52.02/35.15 & 63.36/46.37 & 65.42/48.61 & 61.42/44.31 & 44.27/28.43 & 54.63/37.58 & 19.98/11.10 & 57.30 & 40.58 \\
			TransUNet \cite{chen2021transunet} &R50-ViT-B& 52.00/35.13 & 74.14/58.90 & 72.87/57.57 & \textbf{80.68}/\textbf{67.61} & 41.19/25.93 & 54.03/37.01 & 34.50/20.85 & 64.37 & 49.23 \\
			UNetformer  \cite{wang2022unetformer} &ResNet-18& 54.66/37.61 & 69.09/52.78 & 68.33/51.89 & 77.66/63.47 & \textbf{56.98}/\textbf{39.84} & 51.01/34.23 & 20.54/11.44 & 65.34 & 49.12 \\
			CMTFNet \cite{wu2023cmtfnet} &ResNet-50& 56.09/38.98 & \textbf{74.18}/\textbf{58.96} & 67.11/50.50 & 70.3/54.27 & 47.00/30.72 & 54.45/37.41 & 41.92/25.65 & 62.95 & 46.68 \\
			\hline
			RS3Mamba &R18-Mamba-T& \textbf{56.86}/\textbf{39.72} & 74.02/58.75 & \textbf{73.35}/\textbf{57.92} & 75.78/61.00 & 54.27/37.24 & \textbf{56.80}/\textbf{39.67} & \textbf{50.72}/\textbf{33.98} & \textbf{66.86} & \textbf{50.93} \\
			\hline
	\end{tabular}\label{tab:ulist}
\end{table*}

\begin{figure}[hbp]
\centering
{\includegraphics[width=1.0\linewidth]{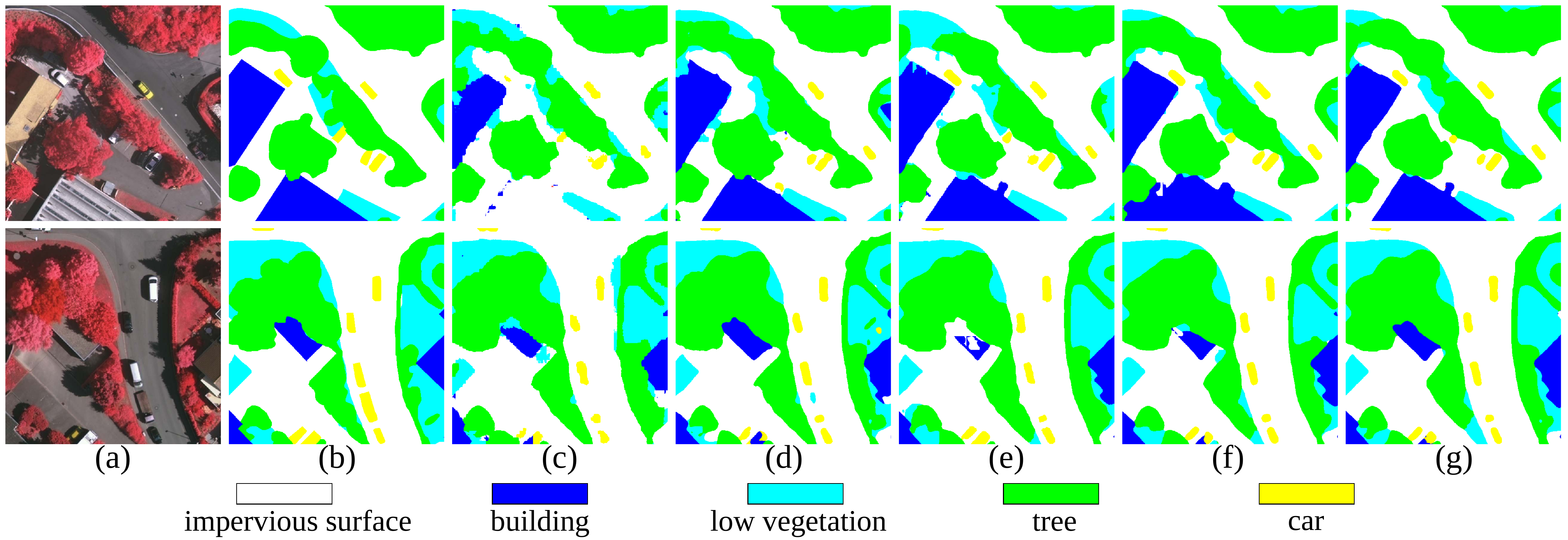}}
\caption{Qualitative performance comparisons on the ISPRS Vaihaigen with the size of $512 \times 512$. (a) NIRRG images, (b) Ground truth, (c) ABCNet, (d) TransUNet, (e) UNetformer, (f) CMTFNet and (g) the proposed RS3Mamba. We showcase two samples for each model.}
\label{fig3}
\end{figure}

\subsection{Performance Comparison}
\subsubsection{Performance Comparison on the Vaihingen dataset}
As shown in Table~\ref{tab:vlist}, the proposed RS3Mamba exhibits great improvements in terms of mF1 and mIoU compared to the baseline UNetformer. This confirms that the dual-branch architecture based on VSS blocks effectively enhances feature extraction. When compared with existing state-of-the-art models, RS3Mamba outperforms in four classes: {\em impervious surface}, {\em low vegetation}, {\em tree} and {\em car}. Notably, RS3Mamba achieves a performance improvement of $0.30\%$ and $0.53\%$ in F1 and IoU, respectively, for the {\em impervious surface} class compared to CMTFNet. Furthermore, the IoU for {\em low vegetation} is enhanced by $0.51\%$, and the IoU for {\em building} is enhanced by $0.44\%$ compared to the baseline UNetformer. In terms of overall performance, RS3Mamba achieves an mF1 score of $90.34\%$ and an mIoU of $82.78\%$, representing increases of $0.49\%$ and $0.81\%$, respectively, compared to the corresponding performance of the baseline UNetformer. These improvements are primarily attributed to the global semantic information provided by the auxiliary branch. Fig.~\ref{fig3} illustrates visualization examples of the results obtained by all five methods under consideration. It is evident that RS3Mamba can more accurately segment ground objects with smoother borders and fewer noise points.

\begin{figure}[t]
\centering
{\includegraphics[width=1.0\linewidth]{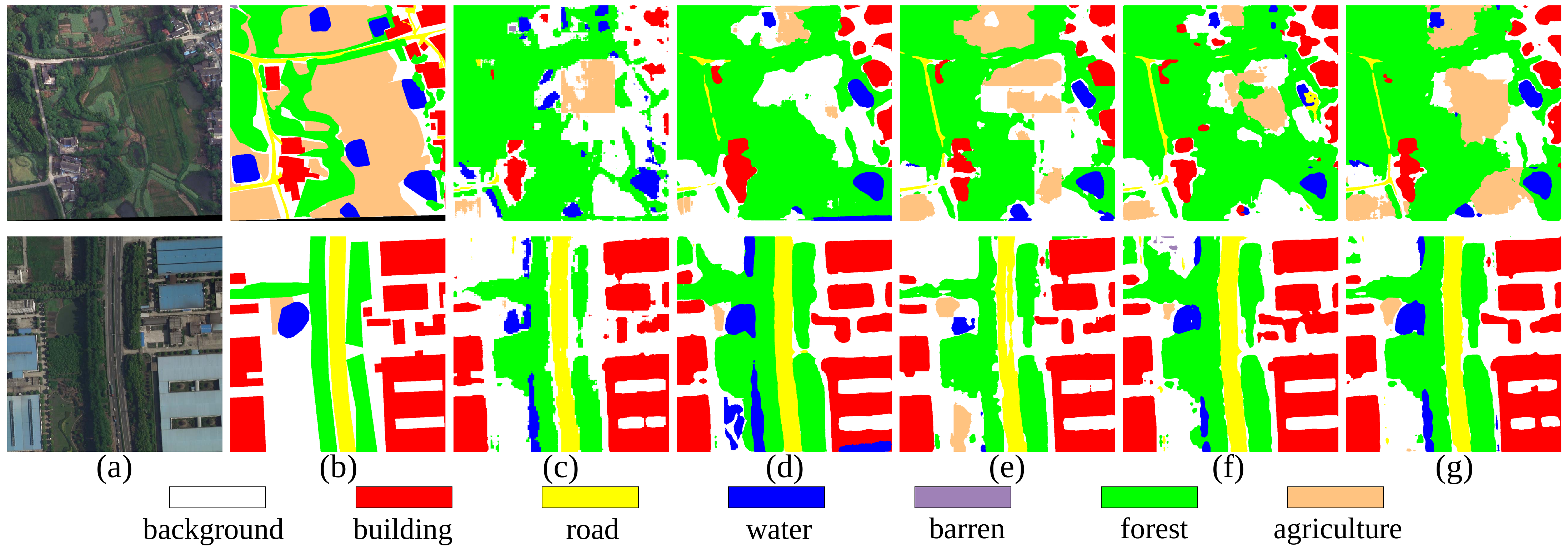}}
\caption{Qualitative performance comparisons on the LoveDA Urban with the size of $1024 \times 1024$. (a) NIRRG images, (b) Ground truth, (c) ABCNet, (d) TransUNet, (e) UNetformer, (f) CMTFNet and (g) the proposed RS3Mamba. We showcase two samples for each model.}
\label{fig4}
\end{figure}

\subsubsection{Performance Comparison on the LoveDA Urban}
Despite differences in sample resolution and ground object categories between the two datasets, the experiments on the LoveDA Urban dataset showed similar results on the ISPRS Vaihingen dataset. As shown in Table~\ref{tab:ulist}, the segmentation performance for {\em background}, {\em road},  {\em forest} and {\em agriculture} were $56.86\%/39.72\%$, $73.35\%/57.92\%$, $56.80\%/39.67\%$, and $50.72\%/33.98\%$, respectively, which amounts to an increase of $0.77\%/0.74\%$, $0.48\%/0.35\%$, $2.17\%/2.09\%$, and $8.8\%/8.33\%$ in F1 and IoU, respectively, as compared to the state-of-the-art methods. In particular, our approach significantly improves the {\em agriculture} category, which is the most difficult to detect among existing methods, proving the potential of RS3Mamba for the recognition of difficult category. In terms of overall performance, RS3Mamba improves over the baseline an mF1 score of $1.52\%$ and an mIoU of $1.81\%$, respectively. Fig. \ref{fig4} presents visualization examples from the LoveDA Urban. The more complete identification for ground objects was apparent, which significantly demonstrates the effectiveness and necessity of global auxiliary information for the representations learning of remote sensing images.

\begin{table}[t]\centering
\begin{threeparttable}
	\centering
	\caption{Ablation studies on the dual-branch and CCM. (\%). Bold values are the best}
	\setlength{\tabcolsep}{5mm}{
		\begin{tabular}{ccc|cc}
			\hline
      \textbf{Res-b} &\textbf{VSS-b} & \textbf{CCM}   &  \textbf{mF1}  & \textbf{mIoU} \\
			\hline
			\checkmark & &   & 89.85 & 81.97  \\
			& \checkmark & & 89.04 & 80.58 \\
			\checkmark& \checkmark & & 90.21 & 82.57 \\
			\hline
		  \checkmark& \checkmark &  \checkmark  & \textbf{90.34} & \textbf{82.78} \\
			\hline
	\end{tabular}}\label{tab:abla}
\end{threeparttable}
\end{table}

\subsection{Ablation Study}
To verify the effectiveness of the proposed VSS auxiliary branch and CCM module inRS3Mamba, four ablation experiments were performed as shown in Table~\ref{tab:abla}. In addition to the decoder, our network consists of three parts in total, including the residual main branch (Res-b), VSS auxiliary branch (VSS-B), and CCM for feature fusion. The first row is our baseline method, named UNetformer, whereas the second row shows the case for the residual main branch being replaced by the VSS auxiliary branch.  It can fully illustrate the current performance of VSS-based encoders on remote sensing images. Furthermore, the third row shows the case in which the dual-branch is adopted but we only use simple element-wise summation for feature fusion instead of CCM.  Inspection of the first two rows in Table~\ref{tab:abla} suggests that the existing VSS-based backbones are still in the development stage, and due to its insufficient pre-training, it is hard to outperform the classic ResNet. Furthermore, the results in the third row illustrate the validity of our proposed dual-branch structure. As an auxiliary branch, VSS encoder can provide sufficient additional information to assist feature extraction and final semantic recovery. The results of the third and fourth rows demonstrate the effectiveness of our CCM, which further excavates the features effectively according to the structure characteristics of the dual-branch encoder. Table~\ref{tab:abla} confirmed that the proposed RS3Mamba is an effective framework to explore the VSS-based methods in remote sensing tasks.

\begin{table}[t]
	\centering
	\caption{Computational complexity analysis measured by two $256 \times 256$ images on a single NVIDIA GeForce RTX 4090 GPU. MIoU values are the results on the ISPRS Vaihingen dataset. Bold values are the best.}
	\setlength{\tabcolsep}{5mm}{
		\begin{tabular}{m{1.4cm}<{\centering}|m{0.6cm}<{\centering}m{0.6cm}<{\centering}m{0.6cm}<{\centering}m{0.6cm}<{\centering}}
			\hline
			\textbf{Model}   & \textbf{FLOPs (G)}  &  \textbf{Parameter (M)} & \textbf{Memory (MB)}  &  \textbf{MIoU(\%)}  \\
			\hline
			ABCNet& 7.81 & 13.67 & \textbf{1008}  & 75.20 \\
			TransUNet& 64.55 & 105.32 & 3122 & 78.16 \\
			UNetformer& \textbf{5.87} & \textbf{11.69} & 1010 & 81.97 \\
			CMTFNet& 17.14 & 30.07 & 1872 & 82.12 \\
			\hline
			RS3Mamba& 31.65 & 43.32 & 2332 & \textbf{82.62} \\
			\hline
	\end{tabular}}\label{tab:scale}
\end{table}

\subsection{Model Complexity Analysis}\label{sec:complexity}
The computational complexity of the RS3Mamba is evaluated using three metrics: floating-point operation count (FLOPs), model parameters, and memory footprint. FLOPs accesses into the model's complexity, while model parameters and memory footprint evaluate the network's scale and memory requirements, respectively. An ideal model would exhibit lower values in FLOPs, model parameters, and memory footprint.

Table~\ref{tab:scale} presents the results of complexity analysis for all compared semantic segmentation models considered in this work. Inspection of Table~\ref{tab:scale} shows that our RS3Mamba introduces additional model complexity compared to the baseline UNetformer. This is due to the incorporation of an additional auxiliary branch and the designation of the specific cross-branch feature fusion module. However, it is noteworthy that our approach exhibits a significant reduction in computational complexity and model scale compared to Transformer-based method TransUNet. These outcomes highlight that the effectiveness of our method as a practical way to introduce Mamba into remote sensing tasks. In the early stages of Mamba's development, our method can provide valuable insights for the future development of Mamba in this field.

\section{Conclusion}\label{sec:con}
In this work, we proposed RS3Mamba that firstly introduce a VSS-based model into remote sensing images semantic segmentation tasks. An auxiliary branch based on VSS blocks is presented to provide additional global awareness information with minimal linear computational complexity. To bridge the distinct features of the two branches, we introduce the CAM module, enhancing the cross-branch features from a global and local perspective, respectively, before they are fused by element-wise addition. Compared with existing methods that directly replace CNNs and Transformers with complete VSS models, our method offers a unique way for integrating Mamba. Experimental evaluations conducted on two distinct remote sensing datasets demonstrate that RS3Mamba outperforms other state-of-the-art semantic segmentation methods based on CNNs and Transformers. We hope that our approach will stimulate further exploration of Mamba applications in remote sensing.

\small
\bibliographystyle{ieeetr}
\bibliography{references}

\begin{thebibliography}{10}

\bibitem{diakogiannis2020resunet}
F.~I. Diakogiannis, F.~Waldner, P.~Caccetta, and C.~Wu, ``{ResUNet-a}: A deep
  learning framework for semantic segmentation of remotely sensed data,'' {\em
  ISPRS Journal of Photogrammetry and Remote Sensing}, vol.~162, pp.~94--114,
  2020.

\bibitem{xu2021efficient}
Z.~Xu, W.~Zhang, T.~Zhang, Z.~Yang, and J.~Li, ``Efficient transformer for
  remote sensing image segmentation,'' {\em Remote Sensing}, vol.~13, no.~18,
  p.~3585, 2021.

\bibitem{ma2024miltulevel}
X.~Ma, X.~Zhang, M.-O. Pun, and M.~Liu, ``A multilevel multimodal fusion
  transformer for remote sensing semantic segmentation,'' {\em IEEE
  Transactions on Geoscience and Remote Sensing}, vol.~62, pp.~1--15, 2024.

\bibitem{gu2023mamba}
A.~Gu and T.~Dao, ``Mamba: Linear-time sequence modeling with selective state
  spaces,'' {\em arXiv preprint arXiv:2312.00752}, 2023.

\bibitem{gu2021efficiently}
A.~Gu, K.~Goel, and C.~R{\'e}, ``Efficiently modeling long sequences with
  structured state spaces,'' {\em arXiv preprint arXiv:2111.00396}, 2021.

\bibitem{zhu2024vision}
L.~Zhu, B.~Liao, Q.~Zhang, X.~Wang, W.~Liu, and X.~Wang, ``Vision mamba:
  Efficient visual representation learning with bidirectional state space
  model,'' {\em arXiv preprint arXiv:2401.09417}, 2024.

\bibitem{liu2024vmamba}
Y.~Liu, Y.~Tian, Y.~Zhao, H.~Yu, L.~Xie, Y.~Wang, Q.~Ye, and Y.~Liu, ``Vmamba:
  Visual state space model,'' {\em arXiv preprint arXiv:2401.10166}, 2024.

\bibitem{he2024pan}
X.~He, K.~Cao, K.~Yan, R.~Li, C.~Xie, J.~Zhang, and M.~Zhou, ``Pan-mamba:
  Effective pan-sharpening with state space model,'' {\em arXiv preprint
  arXiv:2402.12192}, 2024.

\bibitem{chen2024rsmamba}
K.~Chen, B.~Chen, C.~Liu, W.~Li, Z.~Zou, and Z.~Shi, ``Rsmamba: Remote sensing
  image classification with state space model,'' {\em arXiv preprint
  arXiv:2403.19654}, 2024.

\bibitem{wang2022unetformer}
L.~Wang, R.~Li, C.~Zhang, S.~Fang, C.~Duan, X.~Meng, and P.~M. Atkinson,
  ``{UNetFormer}: A {UNet}-like transformer for efficient semantic segmentation
  of remote sensing urban scene imagery,'' {\em ISPRS Journal of Photogrammetry
  and Remote Sensing}, vol.~190, pp.~196--214, 2022.

\bibitem{liu2024swin}
J.~Liu, H.~Yang, H.-Y. Zhou, Y.~Xi, L.~Yu, Y.~Yu, Y.~Liang, G.~Shi, S.~Zhang,
  H.~Zheng, {\em et~al.}, ``Swin-umamba: Mamba-based unet with imagenet-based
  pretraining,'' {\em arXiv preprint arXiv:2402.03302}, 2024.

\bibitem{li2021abcnet}
R.~Li, S.~Zheng, C.~Zhang, C.~Duan, L.~Wang, and P.~M. Atkinson, ``Abcnet:
  Attentive bilateral contextual network for efficient semantic segmentation of
  fine-resolution remotely sensed imagery,'' {\em ISPRS journal of
  photogrammetry and remote sensing}, vol.~181, pp.~84--98, 2021.

\bibitem{chen2021transunet}
J.~Chen, Y.~Lu, Q.~Yu, X.~Luo, E.~Adeli, Y.~Wang, L.~Lu, A.~L. Yuille, and
  Y.~Zhou, ``Transunet: Transformers make strong encoders for medical image
  segmentation,'' {\em arXiv preprint arXiv:2102.04306}, 2021.

\bibitem{wu2023cmtfnet}
H.~Wu, P.~Huang, M.~Zhang, W.~Tang, and X.~Yu, ``{CMTFNet}: {CNN} and
  multiscale transformer fusion network for remote sensing image semantic
  segmentation,'' {\em IEEE Transactions on Geoscience and Remote Sensing},
  2023.

\bibitem{liu2021swin}
Z.~Liu, Y.~Lin, Y.~Cao, H.~Hu, Y.~Wei, Z.~Zhang, S.~Lin, and B.~Guo, ``Swin
  transformer: Hierarchical vision transformer using shifted windows,'' in {\em
  Proceedings of the IEEE/CVF international conference on computer vision},
  pp.~10012--10022, 2021.

\bibitem{wang2021loveda}
J.~Wang, Z.~Zheng, A.~Ma, X.~Lu, and Y.~Zhong, ``Loveda: A remote sensing
  land-cover dataset for domain adaptive semantic segmentation,'' {\em arXiv
  preprint arXiv:2110.08733}, 2021.

\bibitem{he2016deep}
K.~He, X.~Zhang, S.~Ren, and J.~Sun, ``Deep residual learning for image
  recognition,'' in {\em Proceedings of the IEEE conference on computer vision
  and pattern recognition}, pp.~770--778, 2016.

\end{thebibliography}
\end{document}